\pdfoutput=1
\documentclass[two-column]{IEEEtran}
\usepackage{amsmath,amssymb,amsfonts,mathrsfs,bm}
\usepackage{amstext}
\usepackage{upgreek}
\usepackage{multicol}
\usepackage{indentfirst}
\usepackage{graphicx}
\usepackage{paralist}
\usepackage{multirow}
\usepackage{tabularx}
\usepackage[noadjust]{cite}

\usepackage{booktabs}
\usepackage{subfigure}
\usepackage{color}
\usepackage{psfrag}

\IEEEoverridecommandlockouts
\allowdisplaybreaks
\usepackage{algorithm,algorithmic}
\floatname{algorithm}{\small \bf Algorithm}

\usepackage{accents}

\usepackage{hyperref}
\usepackage{breakurl}
\allowdisplaybreaks
\usepackage{tikz}
\newcommand*{\circled}[1]{\lower.7ex\hbox{\tikz\draw (0pt, 0pt)%
		circle (.5em) node {\makebox[1em][c]{\small #1}};}}
\begin{document}
\title{Optimizing Wireless Systems Using Unsupervised and Reinforced-Unsupervised Deep Learning}
\author{
 	\IEEEauthorblockN{Dong Liu, Chengjian Sun, Chenyang Yang, and Lajos Hanzo}\\
\thanks{D. Liu and L. Hanzo are with the University of Southampton, Southampton SO17 1BJ, UK (email: {d.liu, hanzo}@soton.ac.uk). C. Sun and C. Yang are with Beihang University, Beijing 100191, China (e-mail: {sunchengjian, cyyang}@buaa.edu.cn).}
\thanks{This work was supported in part by the National Natural Science Foundation of China (NSFC) under Grant No. 61731002 and 61671036. L. Hanzo would like to acknowledge the financial support of the Engineering and Physical Sciences Research Council projects EP/Noo4558/1, EP/PO34284/1, COALESCE, of the Royal Society's Global Challenges Research Fund Grant as well as of the European Research Council's Advanced Fellow Grant QuantCom.}
}
\maketitle
\vspace{-5mm}
\begin{abstract}
Resource allocation and transceivers in wireless networks are usually designed by solving optimization problems subject to specific constraints, which can be formulated as variable or functional optimization. If the objective and constraint functions of a variable optimization problem can be derived, standard numerical algorithms can be applied for finding the optimal solution, which however incur high computational cost when the dimension of the variable is high. To reduce the on-line computational complexity, learning the optimal solution as a function of the environment's status by deep neural networks (DNNs) is an effective approach. DNNs can be trained under the supervision of optimal solutions, which however, is not applicable to the scenarios without models or for functional optimization where the optimal solutions are hard to obtain. If the objective and constraint functions are unavailable, reinforcement learning can be applied to find the solution of a functional optimization problem, which is however not tailored to optimization problems in wireless networks.
In this article, we introduce unsupervised and reinforced-unsupervised learning frameworks for solving both variable and functional optimization problems without the supervision of the optimal solutions. When the mathematical model of the environment is completely known and the distribution of environment's status is known or unknown, we can invoke unsupervised learning algorithm. When the mathematical model of the environment is incomplete, we introduce reinforced-unsupervised learning algorithms that learn the model by interacting with the environment. Our simulation results confirm the applicability of these learning frameworks by taking a user association problem as an example.
\end{abstract}

\section{Introduction}
Both the transceivers and resource allocation of wireless networks such as power allocation, user association, caching policy, etc. have been designed for decades by solving optimization problems.
These problems can be formulated as variable optimization or functional optimization problems, depending on whether the values of the objective function (OF), the constraint function (CF), if any, and the ``variables"  to be optimized change on a similar timescale.

If they vary on a similar timescale, the problem can be formulated as variable optimization, where the optimization variable is a scalar or a vector. Variable optimization problems are quite common in wireless communications~\cite{YW19}. A typical example is to optimize beamforming vectors of multiple users for maximizing the instantaneous sum-rate subject to the instantaneous transmit power constraint. Another example is to optimize a caching policy that maximizes the average spectral efficiency within  the cache-update interval.

If they vary on quite different timescales, the problem is actually functional optimization. A typical example is to optimize the instantaneous transmit power for maximizing the ergodic channel capacity under an average transmit power constraint. In this problem, the transmit power should adapt to fading channels varying on the timescale of milliseconds, while the OF and CF change on the timescale of seconds. In functional optimization, the optimization ``variable" is a function,  mapping for example from the instantaneous channel gain to the transmit power.
Functional optimization is widely used in optimal control theory~\cite{gregory2018constrained}, but it is less familiar to the wireless community, although the classic water-filling power allocation is found by solving the aforementioned functional optimization problem. Given the increasing importance of cross-layer optimization, say for ultra-reliable and low-latency communications~\cite{she2017cross}, functional optimization is gaining attention in designing wireless networks.

Variable optimization problems are often solved by conventional optimization methods. For continuous variables, convex optimization tools~\cite{boyd2004convex} such as the classic interior-point method, and non-convex optimization tools such as semi-definite relaxation (SDR), have been widely used. For discrete variables, various search methods such as the branch-and-bound and cutting-plane methods have been developed. For the sake of promptly adapting to time-varying environments at an affordable computational cost, the solutions are expected to be found before the environment's status (e.g. the instantaneous or
average channel gains, or content popularity) changes. This can be accomplished if the explicit expression of the solution can be derived as a function of the environment's status. Unfortunately, in many cases, conventional optimization algorithms such as the interior point or gradient descent methods~\cite{boyd2004convex} have to be employed for solving the problems via numerical iteration, which impose a high complexity for high-dimensional problems.

\begin{figure*}[t]
	\centering
	\includegraphics[width=0.7\textwidth]{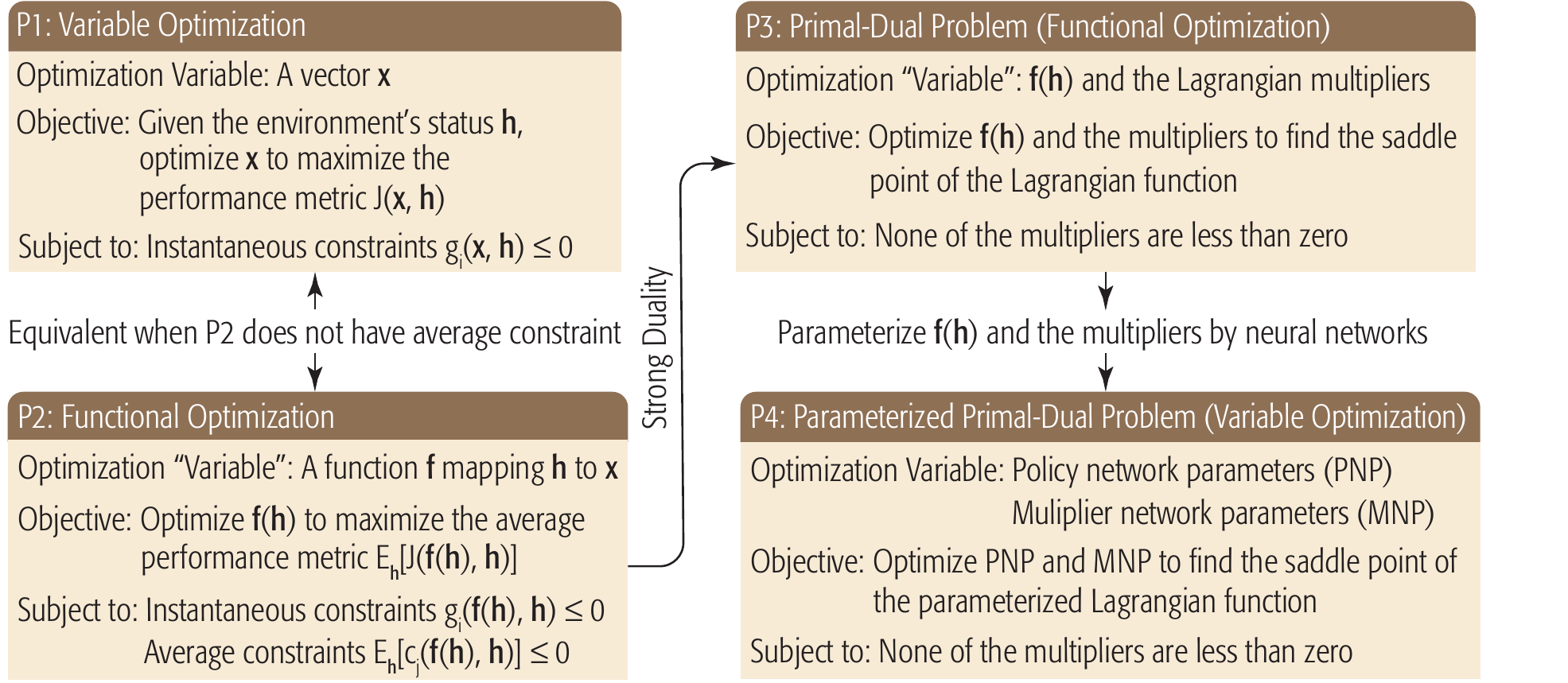}
	\caption{Relationships between variable optimization and functional optimization. It is noteworthy that optimization problems minimizing the OF or having ``$\geq$" or ``$=$" constraints (e.g., minimal data rate constraint) can be easily transformed into problem $\mathsf{P1}$ or $\mathsf{P2}$. }
	\label{fig:equation}
\end{figure*}

Functional optimization problems are in general hard to solve. A conventional technique is the finite element method (FEM)~\cite{Zienkiewicz1977FEM}, which converts functional optimization into high-dimensional variable optimization by only optimizing the values of the function at the sampling points of the environment's status. However, even this procedure may turn out to be prohibitively complex for a large number of sampling points. Consequently, the functional optimization problems of wireless networks are often solved by heuristic techniques~\cite{she2017cross}, which inevitably leads to performance loss~\cite{Chengjian2019GC}.

The holistic optimization problems of next-generation networking become increasingly complex, hence facing the following two challenges:
\begin{itemize}
\item The first one is the computational complexity of their on-line implementation, especially for large-scale physical layer optimization. Due to the rapidly fluctuating nature of fading channels, the optimal solution should be obtained within milliseconds, hence iterative algorithms may require excessive computational resources.

\item The second is the requirement of knowing the environment's model, specifically the expressions of the OF and all the CFs. For conventional optimization algorithms, such as the interior point method, the gradients and even the Hessian matrix of the OF and the CFs with respect to (w.r.t.) the optimization variables are also required. In many scenarios, however, their expressions are unavailable, or it is too complex to derive their gradients.
\end{itemize}

Our ambitious objective is to conceive generic frameworks for solving wireless optimization problems using unsupervised deep neural networks (DNNs). When the environment's model is completely known, or at least the mathematical model (i.e., the OF and the CFs) is known but the distribution of the environment's status is unknown, we conceive an unsupervised algorithm for learning the optimal solution as a function of the environment's status. For the challenging scenarios where the mathematical model is incomplete, we enhance the unsupervised learning algorithm by invoking reinforcements from interactions with the environment for estimating the environment's model.

We commence from recent research efforts invested in handling the aforementioned challenges. We then address model-based unsupervised learning as well as model-free reinforced-unsupervised learning frameworks. Next, we conceive a case-study as a proof-of-concept demonstration of the unsupervised learning frameworks. Finally, we conclude.

\section{State-of-the-Arts}

To reduce the on-line computational complexity, the idea of ``learning to optimize" is proposed for solving variable optimization in~\cite{sun2018learning}. Most recently, a novel framework of using deep learning to find the solution of constrained functional optimization is proposed in~\cite{eisen2019learning}. In fact, reinforcement learning (RL), which is recognized as a powerful tool for model-free problems, is also a generic framework used for solving functional optimization.

\subsection{Variable Optimization Using Supervised and Unsupervised Learning}

Allow us to begin with the standard variable optimization problem $\mathsf{P1}$ of Fig.~\ref{fig:equation}.  Given a vector $\mathbf h$ representing the wireless environment's status (e.g., channel gains of multiple users), the goal is to find a solution $\mathbf x$ (e.g., transmit powers) that maximizes a performance metric $J(\mathbf x, \mathbf h)$ (e.g., sum-rate) subject to some constraints $ g_i(\mathbf x, \mathbf h)\leq 0$ (e.g., maximal transmit power or quality of service constraint).

To reduce the on-line computational complexity of variable optimization problems, the  ``learning to optimize" technique invokes a DNN for approximating the optimized solution as a function of the environment's status~\cite{sun2018learning}. To train the DNN, a training set composed of a number of realizations of the environment's status and the corresponding optimal solution are first generated by running a numerical optimization algorithm for each environment's status. Then, the weights of the neurons in the DNN are optimized using stochastic gradient descent (SGD) for minimizing the empirical mean square error between the outputs of the DNN and the corresponding optimal solution. Once well trained, the DNN can readily approximate the optimal solution for any arbitrary  environment's status, purely requiring computations of forward propagation through a few layers of neurons rather than using conventional optimization algorithms. In this way, the on-line computational complexity is shifted to the off-line supervised training.

Nevertheless, this approach is not applicable to learning general functional optimization in wireless systems, where a function of the environment's status has to be optimized. This is because the conventional algorithms conceived for functional optimization (say the FEM~\cite{Zienkiewicz1977FEM}) suffer from the curse of dimensionality. Even though the off-line cost is less of a concern, the complexity of using FEM to find a numerical solution (i.e., generate one label) increases exponentially with the dimension of the environment's status, while a DNN's training set usually consists of tens of thousands of labels.

Different from the supervised learning approach that requires the optimal solutions under different environment's status as labels for training the DNN, unsupervised learning approach can train the DNN without labels. Very recently, an unsupervised learning approach was conceived for variable optimization problems \cite{YW19}. The basic idea is to employ the OF of the problem as the ``loss" function\footnote{We use the quotation mark on the term ``loss" because the problem may be formulated in the form of  maximizing the OF.} to train the DNN, and hence the labels are no longer necessary.

However, neither of the supervised nor the unsupervised  approaches deal with constraints systematically~\cite{sun2018learning,YW19}. In the supervised learning approach, the constraints are implicitly reflected in the labels, but complex constraints are hard to satisfy by a trained-DNN due to the residual errors between the outputs of DNNs and the feasible solutions. In the unsupervised learning approach of  \cite{YW19}, simple power constraints can be satisfied by selecting an appropriate activation function for the DNN's output layer. However, again, complex constraints cannot be readily satisfied. Furthermore, neither of the two approaches are applicable to problems where the expressions of the OF or CFs are unavailable.

\subsection{Functional Optimization Using Unsupervised Learning and Reinforcement Learning}
To find the solution of constrained functional optimization problems, unsupervised learning techniques are proposed in~\cite{eisen2019learning,Chengjian2019GC,lee2019ICC}. A constrained optimization problem can be transformed into an unconstrained problem by using the Lagrangian approach. The function to be optimized can be parameterized by a DNN, and the DNN can be trained together with the Lagrangian multipliers. This technique was shown to achieve bounded suboptimality under moderate assumptions~\cite{eisen2019learning}. This concept can be used for solving the general functional optimization problem $\mathsf{P2}$ of Fig.~\ref{fig:equation}, which is subject to complex average and instantaneous constraints \cite{Chengjian2019GC,lee2019ICC}.
In \cite{Chengjian2019GC}, it is shown that even the stringent quality of service constraint of ultra-reliable and low-latency communications can be guaranteed. The approaches in~\cite{eisen2019learning,Chengjian2019GC,lee2019ICC} provide more efficient way than FEM in finding numerical solution of functional optimization.

RL aims for finding a function that specifies an action for the state of a system for optimizing a criterion. By appropriately interpreting the three-tuple constituted by $[$\emph{environment's status}, \emph{solution}, \emph{performance metric}$]$ as $[$\emph{state}, \emph{action}, \emph{reward}$]$  in using RL parlance, various RL algorithms can be conceived for learning the relationship between the optimal solution and the environment's status.

RL is eminently suitable for solving problems formulated as Markov decision processes (MDPs), e.g., distributed resource optimization~\cite{calabrese2018learning}, rather than the variable optimization problems formulated as $\mathsf{P1}$ of Fig.~\ref{fig:equation}. In MDP, the current state and action jointly affect the distribution of the next state. Yet $\mathsf{P1}$ is actually a contextual bandit problem (if the elements of $\mathbf x$ are discrete and $\mathsf{P1}$ does not have constraints)~\cite{sutton1998reinforcement}, where the distribution of successor states does not depend on the current action. Therefore, applying RL algorithms to $\mathsf{P1}$ is an overkill. Most RL algorithms use \emph{bootstrapping} for estimating the value of current state or action based on the estimation of the next states' value~\cite{sutton1998reinforcement}, which is however unnecessary for $\mathsf{P1}$ owing to the independence of the next state from the current action. The convergence of RL algorithms that combines bootstrapping, off-policy learning and function approximation (termed as the \emph{deadly triad}~\cite{sutton1998reinforcement}), e.g., the widely adopted deep Q-networks, is not theoretically guaranteed.

Similar to DNNs, RL algorithms have not been specifically designed for optimization problems subject to constraints,
which are however integral parts of most problems in wireless networks. In the literature of wireless communications, the constraints have been treated in a heuristic manner when invoking RL algorithms, e.g., simply bounding the action or adding a term weighted by a pre-determined coefficient in the reward to penalize the violation of constraints. Satisfying the constraints of RL tasks has been investigated in the context of \emph{safe RL}~\cite{garcia2015comprehensive}. The methods proposed along this line of research are designed for the situations where the safety of the agent (say a robot) is particularly important, but only the average constraints are considered. Whether these methods are applicable for optimizing wireless systems remains unclear.

\section{Model-Based Unsupervised Learning}
In this section, we show how to learn the solution of optimization problems as a function of the environment's status using unsupervised DNN, where the model is known or at least partially known. In particular, the expression of the OF and CF is known, while the distribution of environment's status can be unknown. We first introduce how to equivalently transform variable optimization problems to functional optimization problems. Then, we provide a unified framework for both variable optimization and functional optimization.
\begin{figure*}[htb]
	\centering
	\includegraphics[width=0.6\textwidth]{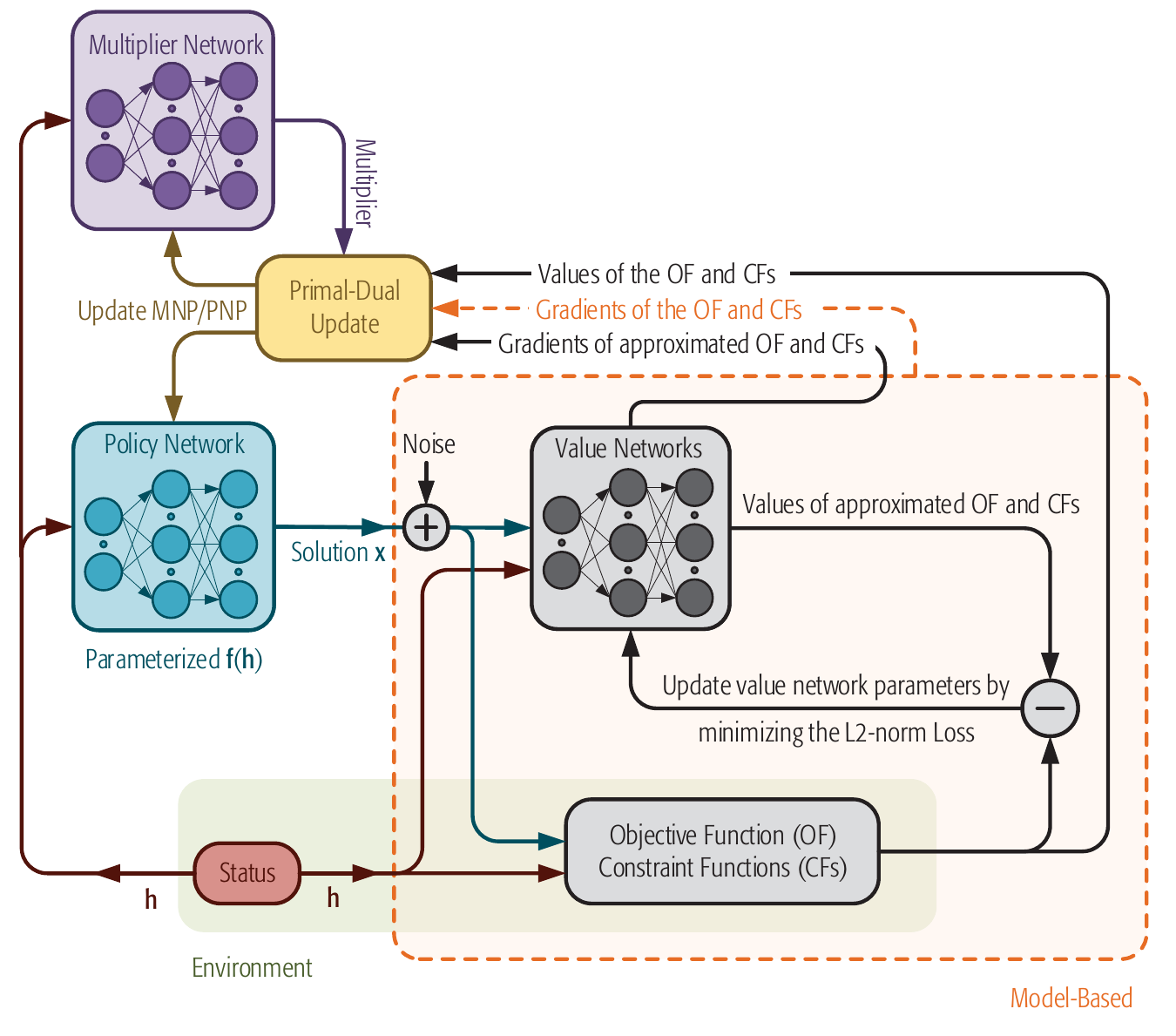}
	\caption{Model-based and model-free unsupervised learning for deterministic policy. The dashed rectangle represents the module required for model-based unsupervised learning, which takes the inputs $\mathbf x$ and $\mathbf h$, and outputs the gradients of the OF and CFs for primal-dual update. The overlapped part between the dashed rectangle and the environment, i.e., the OF and CFs, is the partial knowledge on the environment required for model-based unsupervised learning. In model-free unsupervised learning, the value networks  play the role of approximating the model. }
	\label{fig:det}
\end{figure*}

\subsection{Learning Variable Optimization Without Labels}
Recall that the goal of ``learning to optimize" the variable optimization problem  $\mathsf{P1}$ of Fig.~\ref{fig:equation} is to find a function that maps the environment's status to the corresponding optimal solution. Such a goal can be naturally achieved by functional optimization.

To formulate a functional optimization problem, one needs to find a functional\footnote{A functional is a mapping from functions to scalars, e.g., an integral.} that maps the function to be optimized into a scalar objective to serve as the ``objective function". Furthermore, if we can find a functional to make the resultant functional optimization equivalent to $\mathsf{P1}$, then variable optimization and functional optimization can be learned in a unified framework. As shown in~\cite{Chengjian2019PIMRC}, such a functional can be found as the performance metric averaged over the distribution of the environment's status. Specifically, the relationship between the optimal solution of $\mathsf{P1}$ and the environment's status $\mathbf h$ can be obtained equivalently by finding a function (i.e., a policy) $\mathbf{f}$ that maps $\mathbf h$ to $\mathbf{x}$ (i.e., $\mathbf x = \mathbf f(\mathbf h)$) for maximizing the average performance metric $\mathbb{E}_{\mathbf h}[J(\mathbf f(\mathbf h), \mathbf h)]$. This suggests that we can learn a variable optimization problem by learning its equivalent functional optimization problem, where the optimization ``variable"  is the function $\mathbf{f}(\mathbf h)$.

In problem $\mathsf{P1}$ and its equivalent functional optimization, the constraints are imposed for every realization of the environment's status (i.e., instantaneous constraints). In practice, there may also exist constraints imposed on a function averaged over  the environment's status, e.g., average power constraints or average data rate constraints, as shown in $\mathsf{P2}$. For more general wireless optimization problems, the ``variables'' to be optimized consist of both functions and variables, as exemplified by jointly optimizing a multi-timescale policy~\cite{Chengjian2019GC}.

For convenient exposition, we introduce the unsupervised learning framework for the general functional optimization problem  $\mathsf{P2}$, and we use $\mathbf{f}(\mathbf h)$ and $\mathbf{x}$ interchangeably, since the framework is applicable for both functional and variable optimization. To avoid using labels by repeatedly solving $\mathsf{P2}$, the Lagrangian function\footnote{If an optimization problem does not have any constraint, the Lagrangian function degenerates into the average performance metric.} is used as the ``loss" function for training the DNNs in what follows. In order to compute the gradients of the Lagrangian function, the expressions of the performance metric $J(
\mathbf x, \mathbf h)$, the functions in the instantaneous constraints $g_i(\mathbf x, \mathbf h)$, and the functions in the average constraints $c_j(\mathbf x, \mathbf h)$ have to be available. For brevity, we only refer to $J(\mathbf x, \mathbf h)$ (instead of $\mathbb E_{\mathbf h}[J(\mathbf f(\mathbf h), \mathbf h)]$) as the OF, and only refer to the functions $ g_i(\mathbf x, \mathbf h)$ and $c_j(\mathbf x, \mathbf h)$  (instead of $\mathbb E_{\mathbf h}[c_j(\mathbf f(\mathbf h), \mathbf h)]$) as the CFs.

\subsection{Learning Generic Functional Optimization Without Labels}
To handle the constraints in general functional optimization systematically, $\mathsf{P2}$ is reconsidered in its dual domain. When strong duality holds, $\mathsf{P2}$ is equivalent to finding the saddle point of the Langrangian function~\cite{gregory2018constrained} as in problem $\mathsf{P3}$. This saddle point can be found by optimizing the policy $\mathbf f(\mathbf h)$ and the Lagrangian multipliers for maximizing and minimizing the Langrangian function, respectively. We note that in contrast to the functional optimization considered in~\cite{eisen2019learning}, $\mathsf{P2}$ also takes general instantaneous constraints into consideration whose associated multipliers in $\mathsf{P3}$ of Fig.~\ref{fig:equation} are actually functions of the environment's status~\cite{Chengjian2019GC}.

In general, the functional optimization $\mathsf{P3}$ is hard to solve, because we have to find the optimal value of $\mathbf f(\mathbf h)$ and the multipliers for every possible value of $\mathbf h$.
By resorting to the universal approximation theorem, neural networks can be used for parameterizing the policy and the multipliers in $\mathsf{P3}$, which is more efficient than FEM. In particular, a \emph{policy network} and a \emph{multiplier network} can be employed for parameterizing  the policy~$\mathbf{f}(\mathbf h)$ and the multipliers, respectively. Then, $\mathsf{P3}$ degenerates into a variable optimization problem that optimizes the policy network parameters (PNP) and the multiplier network parameters (MNP) for maximizing and minimizing the Langrangian function, respectively, subject to the constraints that none of the multipliers are less than zero, as shown in $\mathsf{P4}$ of Fig.~\ref{fig:equation}.

If the gradients of the OF and CFs w.r.t. $\mathbf x$ are derivable, then the policy network and multiplier network can be trained by the primal-dual stochastic gradient method that iteratively updates the PNP and MNP along the ascent and descent directions of the Langrangian function's gradients, respectively~\cite{APCC}. The distribution of the environment's status does not have to be known, because the sampled-averaged gradients can be used for replacing the true gradients in the sense of ensemble average. The constraint that none of the multipliers are  less than zero can be satisfied by appropriately selecting the activation function of the multiplier network's output layer, e.g., the \texttt{ReLU}.

The procedure of learning functional optimization is summarized in Fig.~\ref{fig:det}. Upon obtaining a sample (or a batch of samples) of the environment's status $\mathbf h$, the policy network and the multiplier network treat $\mathbf h$ as the input, and then outputs the corresponding solution $\mathbf x$ (i.e., $\mathbf f(\mathbf h)$) and multipliers, respectively. By taking $\mathbf x$ and $\mathbf h$ as the input, the gradients of the OF and CFs are computed based on the environment's model, which is used for training the policy network and multiplier network by primal-dual update. Given sufficient number of the samples for characterizing the distribution of the environment's status and sufficient iterations for the primal-dual updates, the policy network is capable of approximating the relationship between the optimal solution and the environment's status.

\section{Model-Free Reinforced-Unsupervised Learning}
For many problems in wireless networks, there is a lack of complete knowledge concerning the mathematical model, i.e., the expressions of the OF and CFs are unavailable or their gradients cannot be derived analytically. In this section, we introduce model-free unsupervised learning that does not require direct derivation of those gradients.

\begin{figure*}[htb]
	\centering
	\includegraphics[width=0.6\textwidth]{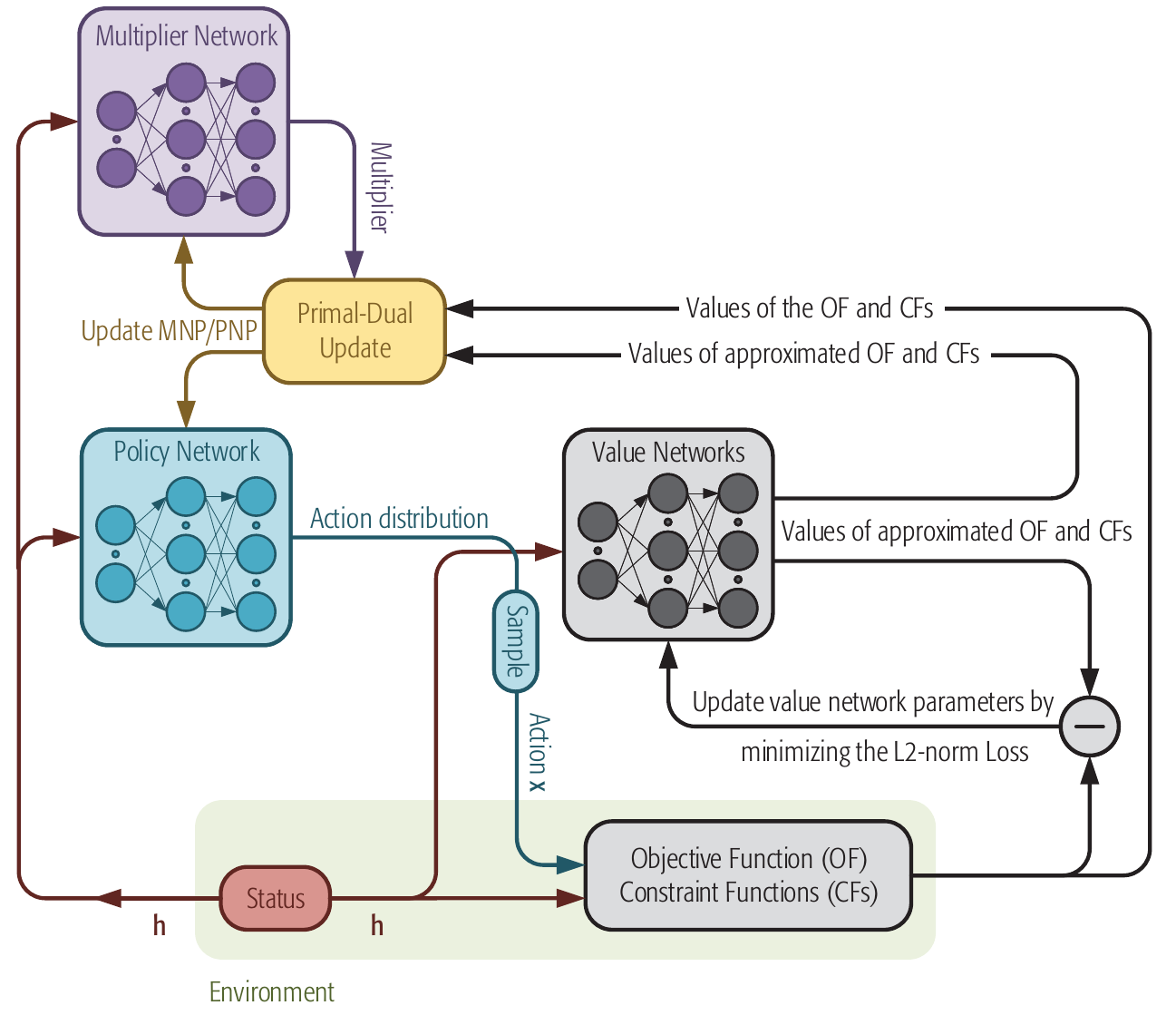}
	\caption{Model-free unsupervised learning for stochastic policy.}
	\label{fig:sto}
\end{figure*}

\subsection{Learning Deterministic Policy}
In practice, the values of the OF and CFs at $(\mathbf x, \mathbf h)$ can be observed after executing $\mathbf x$ (termed as action  in the following) at the environment's status $\mathbf h$. For example, for a power allocation problem, we can quantify the data rate attained, the delay imposed, and the energy consumed after transmitting at power $\mathbf x$ at channel state~$\mathbf h$.

Given the observations of the OF's and CFs' value, the finite difference method can be used for estimating their gradients~\cite{eisen2019learning}, which is however inefficient when these functions have high dimensions and hence they are not applicable to functional optimization problems having infinite dimensions.

Again, based on the universal approximation theorem,  the OF and CFs can be parameterized by neural networks with finite parameters, which are termed as \emph{value networks} in Fig.~\ref{fig:det}. The value networks take $\mathbf x$ and $\mathbf h$ as the input and then they output the approximated values of the OF and CFs at $\mathbf h$. The observed values of the OF and CFs can be used for supervising the training procedure~\cite{APCC}. Specifically, the value networks are trained by minimizing the $L_2$-norm loss between their outputs and the observed values of the OF and CFs using SGD, as shown in Fig.~\ref{fig:det}. Then, the gradients of the approximated OF and CFs are computed via back-propagation of the value networks, which are then used for the primal-dual update of the PNP and MNP.

The employment of the policy network and value network of Fig.~\ref{fig:det} is similar to the \emph{actor-critic} approach of RL. If the functional optimization problem is not subject to constraints, then the gradient of the Langrangian function w.r.t. the PNP degenerates into the \emph{deterministic policy gradient} in RL, which is used for training the actor in the deep deterministic policy gradient (DDPG) algorithm~\cite{DDPG}.

Analogous to the \emph{actor-critic} approach, the policy network, multiplier network, and value networks of Fig.~\ref{fig:det} can be trained simultaneously via interactions with the environment. As shown in Fig.~\ref{fig:det}, upon observing the environment's status, the policy network outputs an action $\mathbf x$ applied to the environment. After observing the values of the OF and CFs evaluated based on the environment, the value network parameters are updated for obtaining a better approximation of the Lagrangian function. Meanwhile, the PNP and MNP are also updated relying on the gradients computed from the value networks for improving the policy. Because the OF and CFs are functions of the action, for accurately approximating their values and gradients at the executed action, it is necessary to obtain their values in the neighborhood of the executed action. For encouraging this exploration, a gradually diminishing noise term is added to the output of the policy network throughout the consecutive iterations.

Since the training procedure does not require the labels generated by finding the numerical solutions of a problem for every environment's status, this approach is still under the framework of unsupervised learning. However, since the unknown OF and CFs are learned via interacting with the environment, we term this framework as reinforced-unsupervised learning.

\subsection{Learning Stochastic Policy}

So far, we have implicitly assumed that the policy to be learned is a continuous function of $\mathbf{h}$, and learned its parameterized form as a deterministic policy in both model-based and model-free unsupervised learning frameworks.
In some scenarios, we aim for finding a discrete policy, e.g., user association, where parameterizing a deterministic policy is not applicable because the output of the neural network is a continuous mapping of its input. Although a discrete policy can be obtained by discretizing a learned deterministic policy, the constraints may not be satisfied after its discretization.

Stochastic policies are widely used in RL for trading off exploration against exploitation~\cite{sutton1998reinforcement}. For functional optimization, the policy network can also parameterize a stochastic policy, which is applicable to both discrete and continuous policy learning~\cite{eisen2019learning}. To realize a stochastic policy, the policy network is designed to take the environment's status as its input and then output the probability distribution of the action to be executed. Then, for each environment's status, the action to be executed is sampled from the distribution given by the output of the policy network, as shown in Fig.~\ref{fig:sto}.

In contrast to the deterministic policy scenario, the sampled gradients of the OF and CFs w.r.t. the executed action are no longer necessary for the primal-dual update in learning a stochastic policy~\cite{APCC}. To train the policy and multiplier networks, the sampled gradients of the Lagrangian function  w.r.t. the PNP and MNP are used for the primal-dual updates, which can be readily obtained via back-propagation~\cite{APCC}.
Nevertheless, the sampled gradient in the primal-dual update of the stochastic policy may exhibit large variance~\cite{sutton1998reinforcement}, because $\mathbf x$ also has to be averaged for obtaining the gradients of the Lagrangian function, which results in slow convergence.

Analogous to the \emph{advantage actor-critic} approach, a {baseline} term that averages the Lagrangian function  over the distribution of the action can be subtracted from the sampled gradient for reducing the variance~\cite{APCC}. To obtain the action-averaged Lagrangian function, the average value of the OF and CFs are again approximated by value networks, which are trained by minimizing the $L_2$-norm loss using SGD, as shown in Fig.~\ref{fig:sto}.

\begin{figure*}[!htb]
	\centering
	\includegraphics[width=0.7\textwidth]{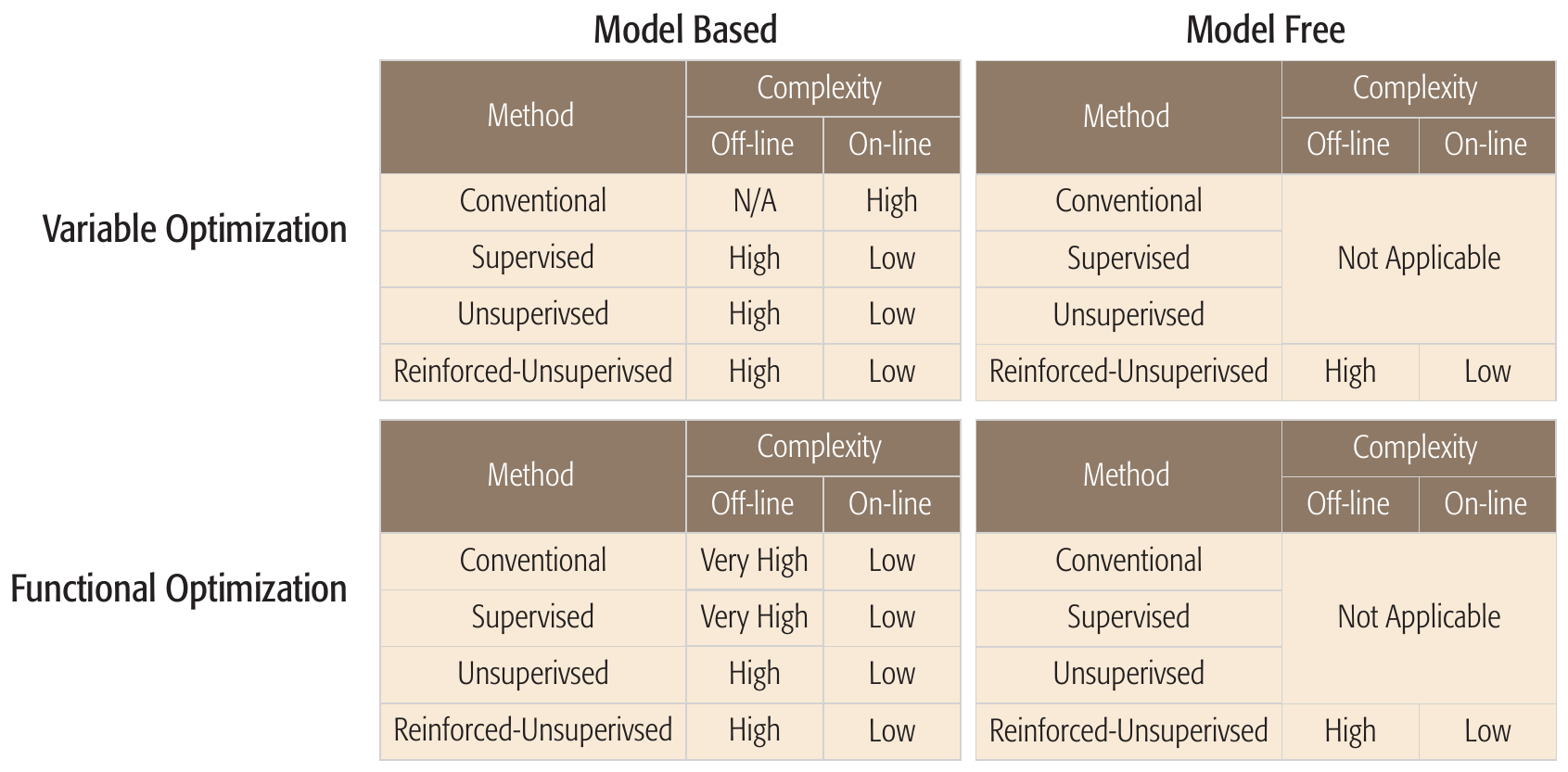}
	\caption{Complexity comparison between the DNN-based frameworks and conventional optimization methods.}
	\label{fig:complexity}
\end{figure*}

The on-line and off-line computational complexity of the DNN-based frameworks and the conventional optimization methods (e.g., convex optimization, SDR, and FEM) are summarized in Fig.~\ref{fig:complexity}.

\section{Case-Study: User Association}
The applicability of model-based and model-free reinforced-unsupervised learning for solving functional optimization has been confirmed by solving a power control problem in \cite{APCC}, where both learning approaches converge to the optimal policy, while satisfying both instantaneous and average constraints. Additionally, the convergence speed of model-free reinforced-unsupervised learning is close to that of model-based unsupervised learning.

In this section, we illustrate the performance of model-based and model-free reinforced-unsupervised learning for solving variable optimization. For convenient interpretation, we consider a simple user association problem optimizing discrete variables. Consider a wireless network supporting $K$ users by $B$ BSs. Each BS can support at most $N$ users and each user can only be associated with one BS. We aim for optimizing the user association based on the received signal-to-noise ratios (SNRs) of every user w.r.t. each BS (which represents the environment's status) to maximize the sum-rate of all the users.

In the simulation, the achieved sum-rate is computed by Shannon's formula, while during the learning process, the relationship between the sum-rate and the SNRs is assumed to be unknown to show the applicability of the model-free unsupervised learning when no model is available. We consider the scenario with two BSs, three users, and each BS can support at most two users. The BSs are positioned along a road with $500$~m inter-BS distance. The minimum distance between the BSs and the road is $100$~m. The users are uniformly distributed along the road. Each user is served with $20$ W transmit power and $10$ MHz bandwidth. 

To learn the relationship between the optimal user association and the SNRs under the user association constraint, we parameterize a stochastic policy. The policy network takes the SNRs as its input and then outputs the probability distribution of user association solution $\mathbf x$. The value network is used for approximating the sum-rate averaged over $\mathbf x$. A multiplier network is used for ensuring that the number of users associated with each BS does not exceed $N$. In general, we can also use the multiplier network for ensuring that the sum of the association probabilities for different BSs as one. However, to satisfy this probability constraint, we can simply use the \texttt{softmax} as the activation function of the policy network's output layer, since such a probability constraint is common for all stochastic policies. The \texttt{ReLU} is used as the activation function of the value and the multiplier networks' output layers to yield positive outputs. We use the \texttt{tanh} as the activation function for the hidden layers of all networks for avoiding the performance loss caused
by gradient vanishing. Each of the neural networks consists of two fully connected layers and $20$ neurons in each layer. The learning rates for all networks are $0.01/(1+0.001 t)$, where $t$ is the number of iterations.

\begin{figure}[!htb]
	\centering	
	\subfigure[Average rate.]{
		\label{fig:UserSchd_Obj} 
		\includegraphics[width=0.45\textwidth]{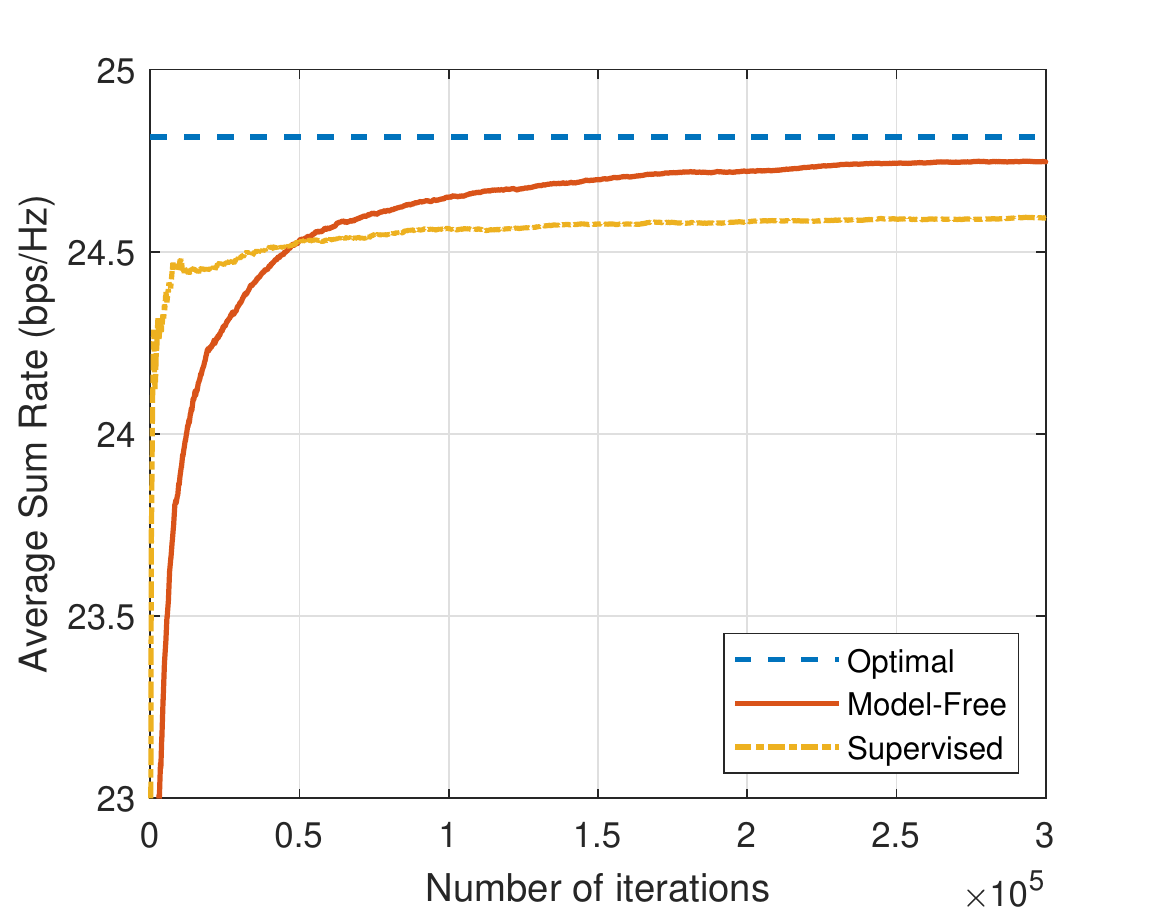}}
	\subfigure[Constraint violations.]{
		\label{fig:UserSchd_Con} 
		\includegraphics[width=0.45\textwidth]{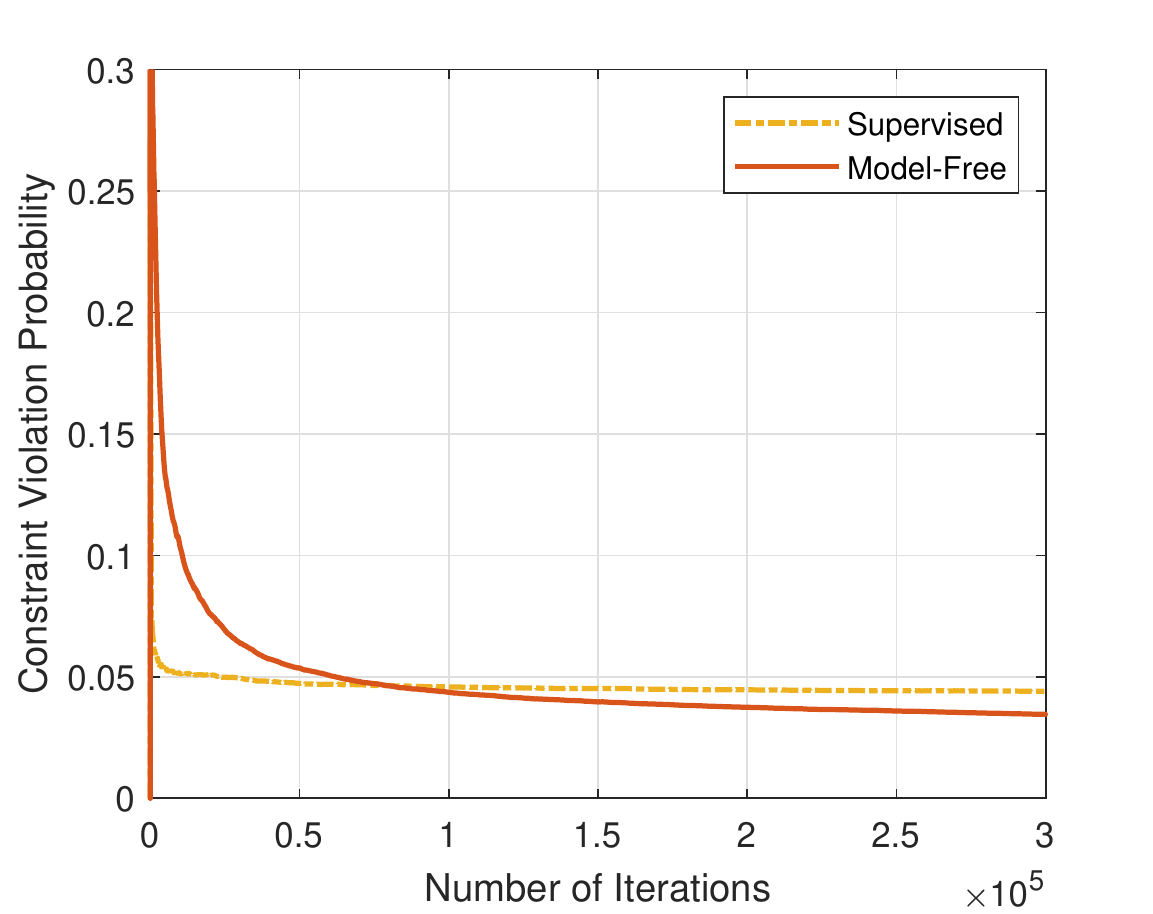}}
	\caption{Convergence of learning the user association policy, where the average is taken over $500$ successive iterations.}
	\label{fig:UserSchd}
\end{figure}

We note that for the stochastic policy, there is no substantial difference between the model-free and the model-based unsupervised learning, since the gradients of the OF and CFs are not required for the primal-dual updates of the policy and multiplier networks. Therefore, we only compare the model-free reinforced-unsupervised learning to the optimal solution obtained via exhaustive search and to the supervised learning that is trained via the supervision of the optimal solutions. The convergence results are provided in Fig.~\ref{fig:UserSchd}. We can see that when the number of iterations is low, the model-free approach experiences lower average sum-rate and higher constraint violation probability than the supervised learning approach. This is because the model-free approach has to learn the model by exploring the environment. Upon increasing the number of iterations, the model-free approach converges to the optimal solution and its performance becomes superior to the supervised learning both in terms of the average sum-rate and constraint satisfaction.

\section{Concluding Remarks}

In this paper, we introduced unified frameworks of model-based unsupervised and of model-free reinforced-unsupervised DNNs for learning to optimize variable optimization and functional optimization subject to instantaneous and average constraints. We illustrated how to apply the frameworks to learn discrete variable optimization with the aid of a user association problem. Our preliminary results show that these frameworks are capable of learning the optimal policy, while satisfying the constraints after convergence. In contrast to RL, these frameworks are tailored to non-MDP problems subject to constraints, which cover the majority of wireless optimization problems. Similar to RL, these frameworks are capable of adapting to the changes of the environment's status even without models. Nevertheless, how to reduce the number of iterations required for attaining convergence and achieve near-optimal performance for  more complex problems deserve further investigations.
\bibliographystyle{IEEEtran}
\bibliography{ref}
\end{document}